\documentclass[conference]{IEEEtran}

\IEEEoverridecommandlockouts
\usepackage{cite}
\usepackage{amsmath,amssymb,amsfonts}
\usepackage{algorithmic}
\usepackage{graphicx}
\usepackage{textcomp}
\usepackage{xcolor}
\usepackage{fancyhdr}
\usepackage[hidelinks]{hyperref}

\def\BibTeX{{\rm B\kern-.05em{\sc i\kern-.025em b}\kern-.08em
    T\kern-.1667em\lower.7ex\hbox{E}\kern-.125emX}}
    
\fancypagestyle{firstpagefooter}{%
  \fancyhf{}
  
  \fancyfoot[R]{}
}

\begin{document}
\title{Cattle Detection Occlusion Problem}

\author{\IEEEauthorblockN{}
\IEEEauthorblockA{\textit{} \\
\textit{}\\
}
\and
\IEEEauthorblockN{Aparna Mendu(ccid - amendu), Vaishnavi Mendu(ccid - mendu), Bhavya Sehgal(ccid - bsehgal1)}
\IEEEauthorblockA{\textit{Department Of Computing Science, Multimedia} \\
\textit{University Of Alberta}\\
}
\and
\IEEEauthorblockN{}
\IEEEauthorblockA{\textit{} \\
\textit{}\\
}
}

\maketitle

\begin{abstract}
The management of cattle over a huge area is still a challenging problem in the farming sector. With evolution in technology, Unmanned aerial vehicles (UAVs) with consumer level digital cameras are becoming a popular alternative to manual animal censuses for livestock estimation since they are less risky and expensive.This paper evaluated and compared the cutting-edge object detection algorithms, YOLOv7,RetinaNet with ResNet50 backbone, RetinaNet with EfficientNet and mask RCNN. It aims to improve the occlusion problem that is to detect hidden cattle from a huge dataset captured by drones using deep learning algorithms for accurate cattle detection. Experimental results showed YOLOv7 was superior with precision of 0.612 when compared to the other two algorithms. The proposed method proved superior to the usual competing algorithms for cow face detection, especially in very difficult cases.
\end{abstract}


\thispagestyle{firstpagefooter}

\section{Background}
Since many farms are expanded over large areas, cattle monitoring is still a challenging task. Ground surveillance is still a very popular method used by the farmers to keep track but it is a time consuming task and requires human resources which is expensive. As an alternative unmanned aerial vehicles (UAV) such as drones could be more efficient, less expensive and quicker for a complete visual survey of the herd as in recent times drones are commonly used for inspection and monitoring for example: smart agriculture and urban planning, object like vehicle detection and counting in aerial images.

\vspace{0.3cm}

The project starts with the drones capturing the data sets with the help of RGB and thermal cameras over the desired area as shown below.\\

\includegraphics[scale=0.5]{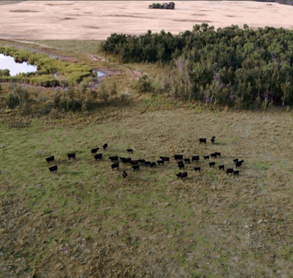}
\includegraphics[scale=0.5]{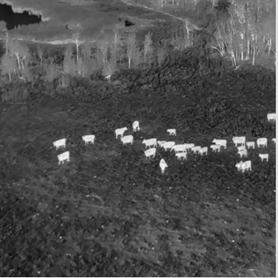}\\
Figure 1: showing images captured by drones and thermal cameras\\

\vspace{0.3cm}

methods have been used for cattle detection as shown in figures below.\\
\vspace{0.3cm}
\includegraphics[scale = 0.5]{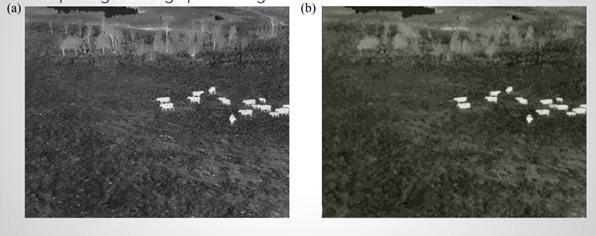}
Figure 1.1: Morphological image processing for cattle detection\\

\vspace{0.3cm}
\includegraphics[scale=0.5]{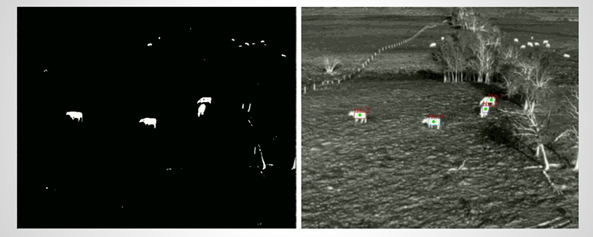}
Figure 1.2: vision based methods used for cattle detection\\

\vspace{0.3cm}
\includegraphics[scale=0.5]{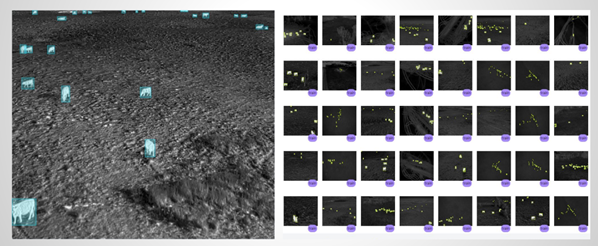}
Figure 1.3: shows the data annotation for cattle detection dataset\\

Below is the convolutional neural network methods used for
detecting objects.

\vspace{0.3cm}
\includegraphics[scale=0.5]{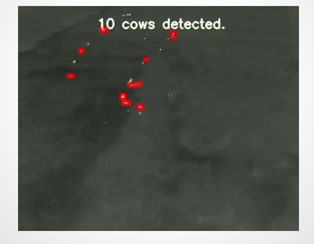}

Figure 1.4: showing results of CNN methods\\

As seen in the results, there still exists issues where hidden cattle are not being detected by these methods, so this project mainly focuses on improving the occlusion problem.\\

\includegraphics[scale = 0.5]{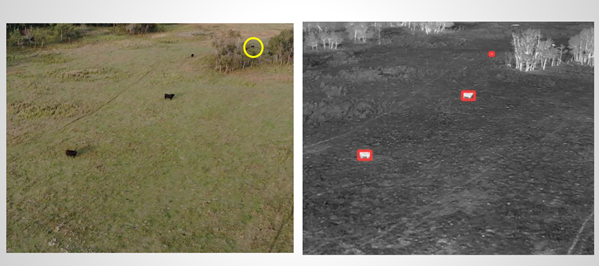}
Figure 1.5: showing the occlusion problem\\

This paper aims to use four deep-learning object detection models, namely, YOLOv7,RetinaNet with ResNet50 backbone,RetinaNet with EfficientNet backbone and Mask RCNN.

\vspace{0.3cm}

YOLOv7 surpasses all known object detectors in both speed and accuracy in the range from 5 FPS to 160 FPS and has the highest accuracy. They propose Extended-ELAN (E-ELAN) based on ELAN for the YOLOv7 model. E-ELAN uses expand, shuffle, merge cardinality to achieve the ability to continuously enhance the learning ability of the network without destroying the original gradient path. They have also analyzed the model re-parameterization strategies applicable to layers in different networks with the concept of gradient propagation path, and propose planned re-parameterized model. In addition, with the dynamic label assignment technology, the training of model with multiple output layers will generate new issues. For this problem, they propose a new label assignment method called coarse-to-fine lead guided label assignment.

\vspace{0.3cm}
\includegraphics[scale = 0.48]{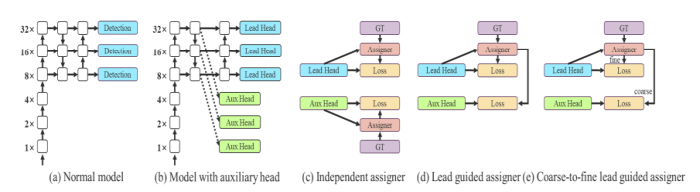}
Figure 2: showing YOLOv7 architecture\\

One of the top one-stage object detection models, RetinaNet has proven successful with small, dense objects.RetinaNet is a one-stage object detection model that addresses class imbalance during training by using a focal loss function.RetinaNet incorporates FPN and adds classification and regression subnetworks to create an object detection model.FPN integrates low-resolution semantically strong features with high-resolution semantically weak features to produce an architecture with rich semantics at all levels. This is accomplished by building a top-down pathway with lateral connections to bottom-up convolutional layers.

\vspace{0.3cm}
\includegraphics[scale = 0.5]{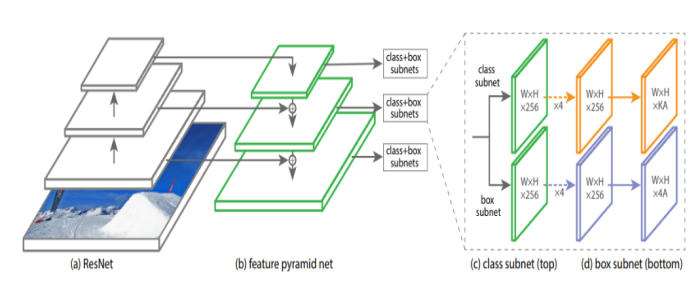}
Figure 2.1: showing RetinaNet architecture\\

EfficientNet is a convolutional neural network architecture and scaling technique that uses a compound coefficient to consistently scale all depth, breadth, and resolution dimensions.A fixed set of scaling coefficients is used by the EfficientNet scaling approach to scale network breadth, depth, and resolution consistently.The foundational EfficientNet-B0 network is built on the squeeze-and-excitation blocks as well as the inverted bottleneck residual blocks of MobileNetV2.

\vspace{0.3cm}
\includegraphics[scale = 0.5]{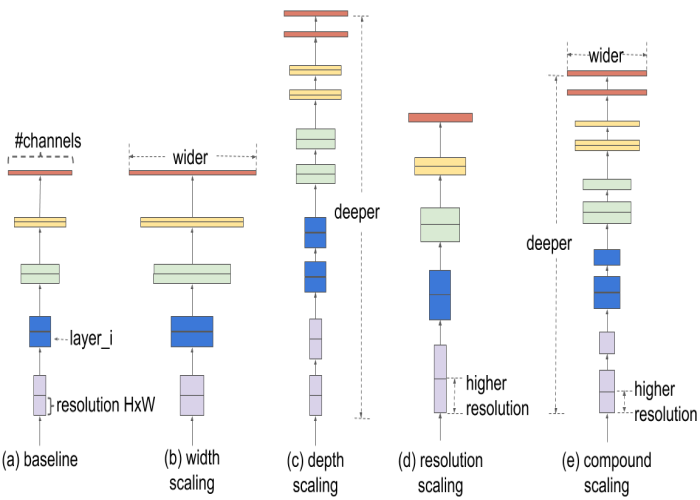}
Figure 2.2: showing EfficientNet architecture\\

Mask R-CNN, is a Convolutional Neural Network (CNN) and state-of-the-art in terms of image segmentation and instance segmentation. Mask R-CNN was built using Faster R-CNN. While Faster R-CNN has 2 outputs for each candidate object, a class label and a bounding-box offset, Mask R-CNN is the addition of a third branch that outputs the object mask. The additional mask output is distinct from the class and box outputs, requiring the extraction of a much finer spatial layout of an object. Mask R-CNN outperforms all existing, single-model entries on every task and is simple to train. Mask R-CNN uses backbone ResNet50.A backbone is a known network trained in many other tasks before and demonstrates its effectiveness.The architectural element which defines how these these layers are arranged in the encoder network and they determine how the decoder network should be built. Few of the Backbone which are widely used are VGG, ResNet, Inception, EfficientNet etc.ResNet backbone was used in Mask R-CNN. A residual neural network (ResNet) is an artificial neural network (ANN). It is a gateless or open-gated variant of the HighwayNet, the first working very deep feedforward neural network with 
hundreds of layers, much deeper than previous neural networks. ResNet architecture is a type of artificial neural network that allows the model to skip layers without affecting performance. 

\vspace{0.4cm}

\includegraphics[scale = 0.5]{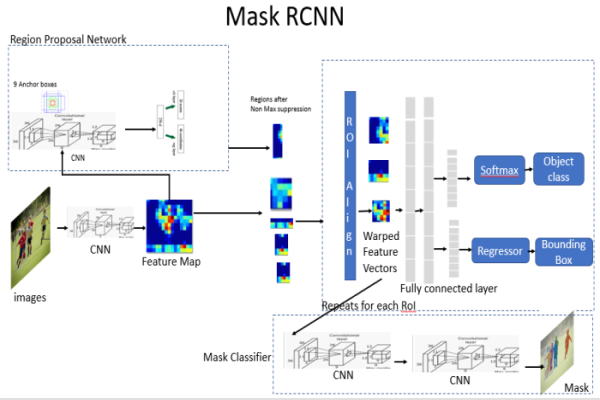}
Figure 2.3: showing Mask RCNN architecture\\
\vspace{0.3cm}

\includegraphics[scale = 0.45]{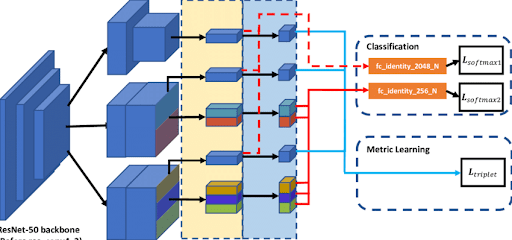}
Figure 2.4: showing ResNet Backbone architecture\\

\section{Related Works}
\subsection{Cattle Detection Using Oblique UAV Images}
\label{sec:oblique}
\vspace{0.1cm}

The imaged dataset used are located at Canchim farm, Brazil and were captured at an altitude of 30 m with respect to the take-off position, using a DJI Mavic 2 Pro equipped with a 20-MPixel camera. The objective of this study was to investigate the feasibility of using a tilted angle to increase the area covered by each image. Deep Convolutional Neural Networks (Xception architecture) were used to generate the models for animal detection. Experimental results show that it is feasible to employ oblique images for detection of animals located up to 250 m from the sensor. One major practical limitation is that most UAVs do not have enough autonomy to cover entire farms in a single flight. This is a problem because, in the time interval between flights, animals may move, weather conditions may vary, and the angle of light incidence will change.

\subsection{Detecting mammals in UAV images: Best practices to address a substantially imbalanced dataset with deep learning.}
\label{sec:imbalance}
\vspace{0.1cm}

Dataset used is acquired over the Kuzikus wildlife reserve in eastern Namibia. This proposed and discussed a methodology for animal censuses based on a Convolutional Neural Network (CNN) and showed how to train deep animal detectors on a real-world UAV image dataset consisting mostly of images with no animals. The results show that the CNN not only manages to yield a substantially higher precision at high recall values when compared to a state-of-the-art detector, but also manages to produce more confined predictions, spreading across a lower number of image tiles. Paper recommend model-agnostic and straight-forward training models to apply to any deep learning-based object detector, and the two evaluation protocols complement the model assessment.

\vspace{0.3cm}
\includegraphics[scale =0.45]{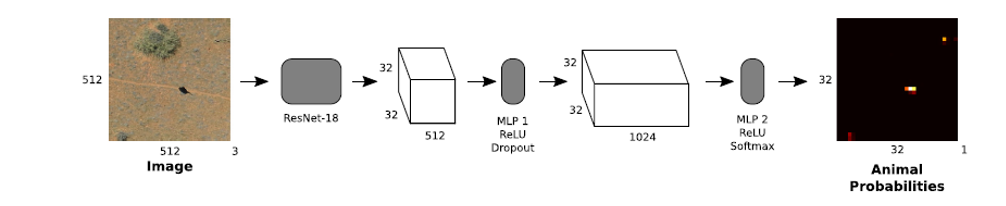}

\subsection{Evaluation of Deep Learning for Automatic Multi-View Face Detection in Cattle}
\label{sec:multiview}
\vspace{0.1cm}

This paper uses RetinaNaet with different different pretrained CNN models(ResNet 50, ResNet 101, ResNet 152, VGG 16, VGG 19, Densenet 121 and Densenet 169) which were then finetuned
by transfer learning and re-trained on the dataset in the paper.Out of all, RetinanNet incorporating ResNet50 was most acuuracy with accuracy of 99.87 percent and average processing time of 0.0438 s. RetinaNet uses Feature Pyramid Networks for feature extraction and focal loss to deal with class imbalance.The results presented indicate that the proposed model was particularly effective for the detection of cattle faces with illumination changes, overlapping, and occlusion. Future work will focus on a lightweight neural network
to improve the running speed of cattle face detection. In addition, future work will also concentrate on building an autonomous livestock individual identification system using
facial features.

\vspace{0.3cm}
\includegraphics[scale = 0.45]{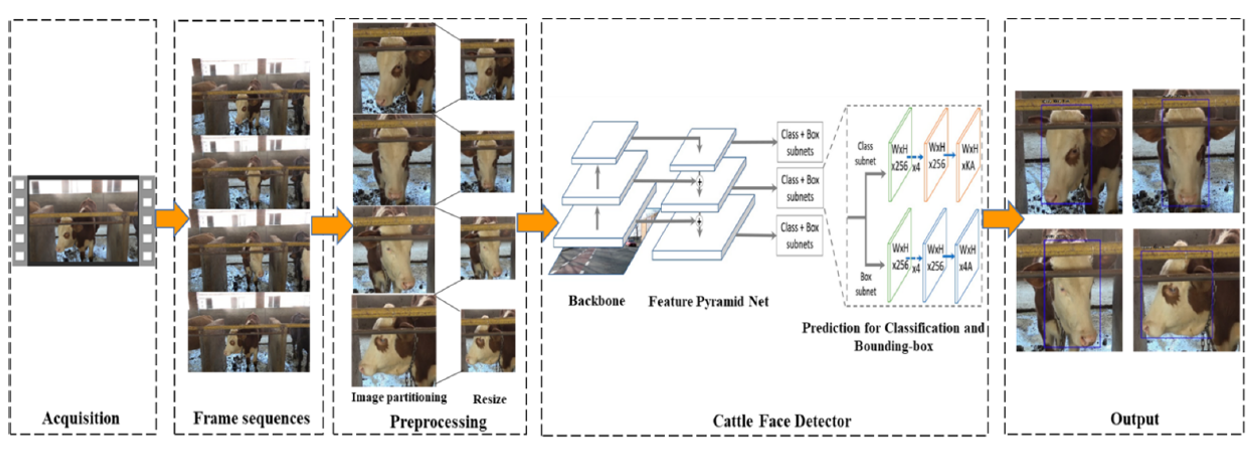}

\subsection{Adapting astronomical source detection software to help detect animals in thermal images obtained by unmanned aerial systems}
\label{sec:astronomical}
\vspace{0.1cm}

This paper used unmanned aerial system equipped with a thermal-infrared camera and software pipeline that we have developed to monitor animal populations for conservation purposes. They have taken multi-disciplinary approach to tackle this problem, we use freely available astronomical source detection software and the associated expertise of astronomers, to efficiently and reliably detect humans and animals in aerial thermal-infrared footage. Then they combined astronomical detection software with existing machine learning algorithms into a single, automated, end-to-end pipeline and test the software using aerial video footage taken in a controlled, field-like environment. This demonstrated that the pipeline works
well and described how it can be used to estimate the completeness of different observational datasets for objects of a given type as a function of height, observing
conditions etc. This paper plans to taking the steps necessary to adapt the system for work in the field and hope to begin systematic monitoring of endangered species in the near future.
\vspace{0.3cm}
\includegraphics[scale = 0.45]{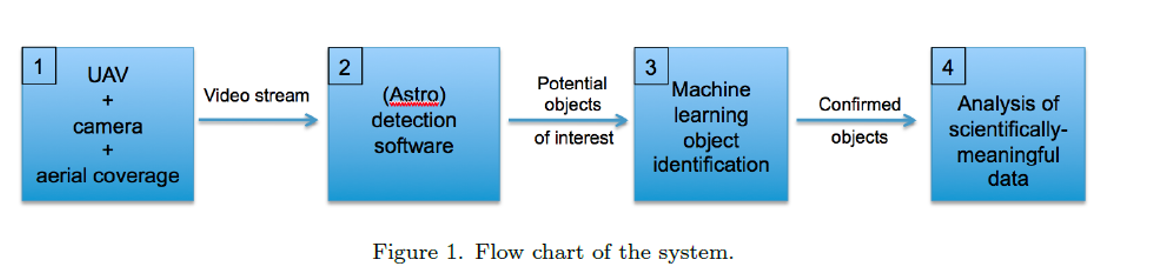}

\subsection{ZoomCount: A Zooming Mechanism for Crowd Counting in Static Images}
\label{sec:zoomCount}
\vspace{0.1cm}

This paper proposes a novel approach for crowd counting in low to high density scenarios in static images. They proposed a simple yet effective modular approach, where an input image is first subdivided into fixedsize patches(224 × 224 size patches) and then fed to a four-way classification module labeling each image patch as low, medium, high-dense or nocrowd. This module also provides a count for each label, which is then analyzed via a specifically devised novel decision module to decide whether the image belongs to any of the two extreme cases (very low or very high density) or a normal case. The specified images pass through dedicated zooming or normal patch-making blocks respectively before routing to the regressor in the form of fixed-size patches for crowd estimate. Even without using any density maps, ZoomCount( RSEpl, RF DBpl) outperforms the state-of-the-art approaches on four benchmark datasets, thus proving the effectiveness of the proposed model.

\vspace{0.3cm}
\includegraphics[scale = 0.45]{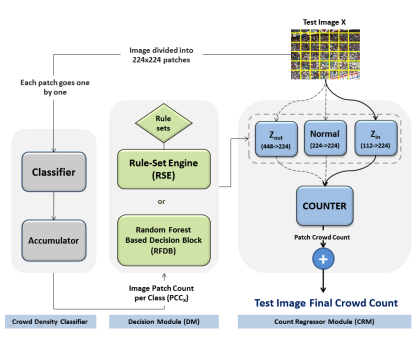}

\subsection{PCC Net: Perspective Crowd Counting via Spatial Convolutional Network}
\label{sec:PCC}
\vspace{0.1cm}

Crowd counting from a single image is a challenging task due to high appearance similarity, perspective changes, and severe congestion. Many methods only focus on the local appearance features and they cannot handle the aforementioned challenges. To tackle this problem, they proposed a perspective crowd counting network (PCC Net). Density map estimation (DME) - learns very local features of density map estimation along with random high-level density classification (R-HDC) extracts global features to predict the coarse density labels of random patches in the image, fore/background segmentation (FBS) encodes mid-level features to segment the foreground and background and Down, Up, Left, and Right (DULR)  module is also used which is based on spatial convolutional networks to encode the perspective changes. Extensive experiments demonstrate that the proposed PCC Net achieves competitive results. Especially, PCC Net significantly improves the counting performance for extremely congested crowd scenes. 

\vspace{0.3cm}
\includegraphics[scale = 0.45]{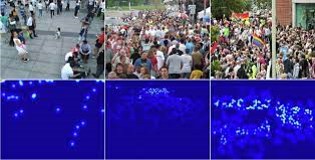}

\subsection{Detection, Tracking, and Counting Meets Drones in Crowds: A Benchmark}
\label{sec:benchmark}
\vspace{0.1cm}

In this paper the authors have proposed developments in object detection, tracking and counting algorithms for crowd counting. They have collected a large-scale drone-based dataset for density map estimation, crowd localization and tracking namely DroneCrowd which consists of 112 video clips. These video clips are annotated with more than 4.8 million head annotations and several video-level attributes. They propose the STNNet method to jointly solve density map estimation, localization, and tracking in drone-captured crowded scenes. STNNet is formed by four modules, i.e., the feature extraction subnetwork, followed by the density map estimation heads, the localization, and the association subnets. To exploit the temporal consistency, the association subnet is designed to predict motion offsets of targets in consecutive frames for tracking. Besides, they developed the neighboring context loss to guide the training of association subnet. The STNNet performs favorably against the state-of-the-art methods in the overall testing set. It indicates that the proposed method generates more accurate and robust density maps in different scenarios.

\vspace{0.3cm}
\includegraphics[scale = 0.45]{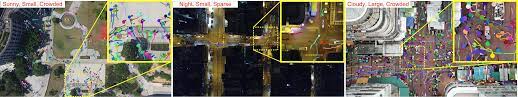}

\subsection{Dynamic Momentum Adaptation for Zero-Shot Cross-Domain Crowd Counting}
\label{sec:zero}
\vspace{0.1cm}

Online fine-tuning is time-consuming. Moreover, in real-world applications, labeled data in the target domain may not always be available for fine-tuning, and large-scale unlabeled data in the target domain may also be unavailable (e.g., for a new camera installation). existing counting methods suffer a large performance drop in this zero-shot cross-domain evaluation, due to the lack of online adaptation to bridge the domain gap.
Crowd Counting framework built upon an external Momentum Template termed C2MoT, which enables the encoding of domain-specific information via an external template representation. The Momentum Template (MoT) is learned in a momentum updating way during offline training, and then is dynamically updated for each test image in online cross-dataset evaluation. Since it dynamically updated MoT,  the C2MoT effectively generates dense target correspondences that explicitly accounts for head regions, and then effectively predicts the density map based on the normalized correspondence map.

\includegraphics[scale = 0.6]{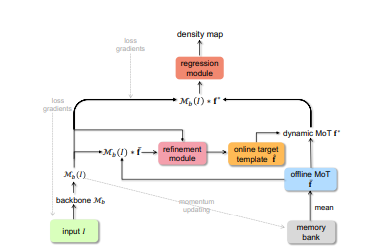}

\subsection{To Choose or to Fuse? Scale Selection for Crowd Counting}
\label{sec:choose}
\vspace{0.1cm}

In this paper, we address the large-scale variation problem in crowd counting by taking full advantage of the multiscale feature representations in a multi-level network. They propose a Scale-Adaptive Selection Network (SASNet), which automatically learns the internal correspondence between the scales and the feature levels. Instead of directly using the predictions from the most appropriate feature level as the final estimation, The SASNet also considers the predictions from other feature levels via weighted average, which helps to mitigate the gap between discrete feature levels and continuous scale variation. The PRA Loss acts as a fine-grained complement for the patch-wise feature level selection strategy, and helps to reduce the inconsistency between training target and evaluation metric. After extensive experiments on the four challenging datasets have proved the effectiveness of the contributions.

\vspace{0.3cm}
\includegraphics[scale = 0.45]{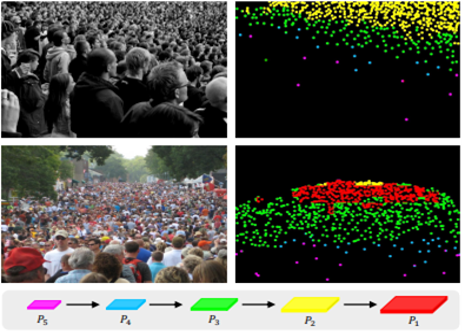}

\subsection{Enhancement of low visibility aerial images using histogram truncation and an explicit Retinex representation for balancing contrast and color consistency}
\label{sec:Retinex}
\vspace{0.1cm}

This paper aims to enhance images that can assist pilots flying under poor visibility and improves on existing multi-scale Retinex with color restoration algorithms. It applies new strategy to construct multiple scales, derive a novel explicit formulation to balance contrast and color constancy, and introduce a histogram truncation approach to obtain improved results. Future work of paper aims to conduct experiments on more outdoor images, consider optimizing the algorithm to put it into wide use and use parallel implementation to attain realtime performance. 

\subsection{A multi-scale attentive recurrent network for image dehazing}
\label{sec:dehaze}
\vspace{0.1cm}

A multi-scale attentive recurrent network is proposed for image dehazing, which consists of a haze attention map predicted network and a recurrent encoder-decoder network. The proposed network is implemented using the PyTorch framework and is trained on a PC with 4 NVIDIA RTX 2080 Ti GPUs. The input images are resized to 512×512 with a batch size of two. For loss optimization, the Adam optimizer is utilized. The kernel sizes are set to 3 and the strides are set to 2, for all the convolutional layers and deconvolutional layers in the network. This method outperforms existing methods in terms of visual and quantitative results , the time complexity is also reduced and the average run time for dealing with a hazy image is 0.039s. Like most proposed image dehazing methods, this method also does not work for images taken in the evening with artificial light; this is due to the training dataset not containing similar scenes. In future, to satisfy the practical requirements we can enrich the training dataset and improve the robustness of the framework.

\vspace{0.3cm}
\includegraphics[scale = 0.5]{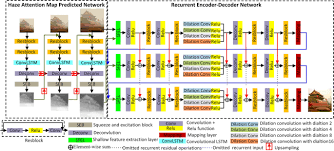}

\subsection{Fog Removal of Aerial Image Based on Gamma Correction and Guided Filtering}
\label{sec:fog}
\vspace{0.1cm}

In order to improve the visual effect of aerial images, a combination of Gamma correction and Retinex defogging algorithm is proposed for aerial foggy images. In this paper, according to the Retinex, an image dehazing enhancement algorithm, the details are lost, the color is distorted, and so on. Before the image is defogged, Gamma correction is performed to correct the contrast and brightness of the original image, and then the corrected images are subjected to three-scale guided filtering processing. The filtered image is weighted and fused, and then introduced in the single scale Retinex model to defog aerial images. Finally, Weighted fusion is implemented to finally enhance aerial images. The proposed method in this paper achieves high contrast and high color reproduction. Compared to the dark channel dehazing and standard MSRCR algorithm.

\subsection{Cascaded Region Proposal Networks for Proposal-Based Tracking}
\label{sec:cascade}
\vspace{0.1cm}

This paper proposes a cascaded region proposal network framework for visual tracking based on region proposal networks, spatial transformer networks and proposal selection strategy. They designed the structure of cascaded region proposals networks and offer the training method and proposal selection method based on the network, so as to obtain high-quality samples, secondly, They propose the feature extraction model based on spatial transformer network to improve the robustness of tracking method for spatial rotation and scale transformation then we utilize the proposals selection and re-ranking strategy to remove the easy anchor through the method of center localization filter and scale change penalty, and effectively estimate the localization and scale of the tracked target. The proposal selection strategy guarantees to provide the high-quality anchors for tracking procedure based on scale change penalty which improves the tracking accuracy and speed.

\vspace{0.3cm}
\includegraphics[scale = 0.5]{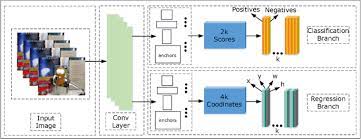}

\subsection{Synthetic Aerial Image Generation and Runway Segmentation}
\label{sec:Runway}
\vspace{0.1cm}

Vision assisted navigation is an active area of research to assist pilots during bad weather conditions but most of the existing systems are not accurate. The proposed approach relies only on the availability of the GPS+AHRS data and the 3D model and produces output with a good visual quality and accuracy comparable to the actual images captured by the aircraft. A synthesized image sequence obtained from a 3D model of a view would be effective in conveying 3-D characteristics of the vanishing point (intersection between the horizon line and runway axis) and the beginning of the runway to the pilot. This can help to improve the pilot’s visual perception of the surroundings under adverse weather conditions, leading to a safer landing.

\subsection{Poisson Surface Reconstruction from LIDAR for Buttress Root Volume Estimation}
\label{sec:LIDAR}
\vspace{0.1cm}

Tree buttress volumes are significant in analyzing bioinformation of a forest. However, forestry researchers have no accurate method for estimating the volume of complex tree buttress roots. The proposed new method with the following components: (1) scanning and collecting point cloud data; (2) computing normals of the point clouds; (3) applying the Poisson Surface Reconstruction algorithm to the point clouds; (4) segmenting the buttress part and closing holes; and, (5) calculating the mesh volume. The advantage of this method over other volume estimation methods is that it has high accuracy for buttresses of different shapes. 

\subsection{Background Subtraction based on Principal Motion for a Freely Moving Camera}
\label{sec:freely}
\vspace{0.1cm}

They have used videos from the Freiburg-Berkeley Motion Segmentation Dataset (FBMS-59) is utilized for evaluation. They used the bear02, meerkats01 and lion02 videos for the evaluation experiments. This paper analyze the principal motion of pixels for subtracting background in videos obtained from freely moving cameras. Robust Principal Components Analysis (RPCA) is utilized for subtracting the background of videos obtained from freely moving cameras, in which both angle and magnitude of the optical flow are utilized for the analysis of motion and super-pixels are utilized to compensate for the defects produced by inaccuracies in optical flow, where a double threshold strategy is proposed to integrate the foreground results captured from the angle as well as the magnitude. Although the proposed approach did not solve the problem perfectly, it proposed a new method that can approach the solution in a new way. 

\vspace{0.3cm}
\includegraphics[scale = 0.3]{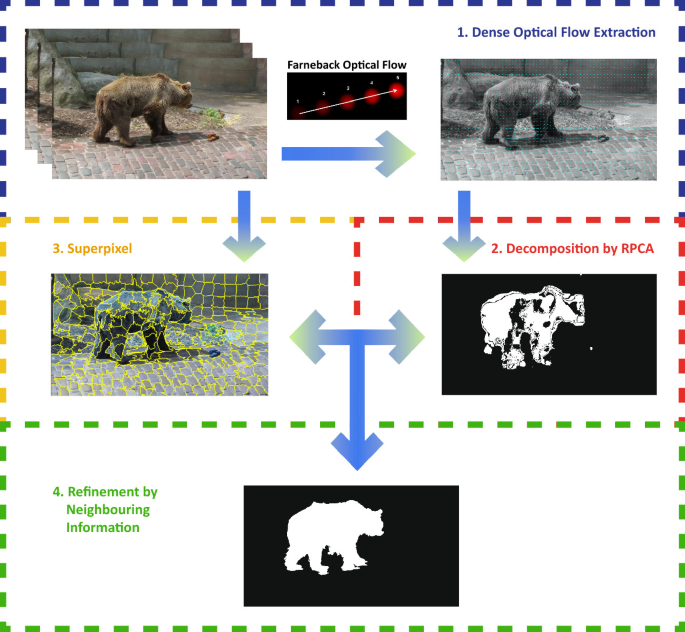}

\subsection{Vehicle Detection and Counting in High-Resolution Aerial Images Using Convolutional Regression Neural Network }
\label{sec:vehicle}
The proposed system has been evaluated on two public datasets namely DLR Munich vehicle dataset provided by Remote Sensing Technology Institute of the German Aerospace Center and Overhead Imagery Research Data Set (OIRDS) dataset. The algorithms for vehicle detection in the literature can be categorized into two groups: shallow-learning-based methods and deep-learning-based methods. It used regression model in order to predict the density map of the input patches. Then, the output of FCRN goes under empirical threshold which results a binary image. Finally, a simple connected component algorithm is used for ending the locations and count of the blobs that represent the detected vehicles. They achieved the highest true positive rate and the lowest false alarm rate. the F1 and precision scores are better than the state-of-the-art methods. The proposed system consumes more time during inference compared with the other systems. The future work will be focusing on a much faster model with better performance.

\vspace{0.3cm}
\includegraphics[scale = 0.45]{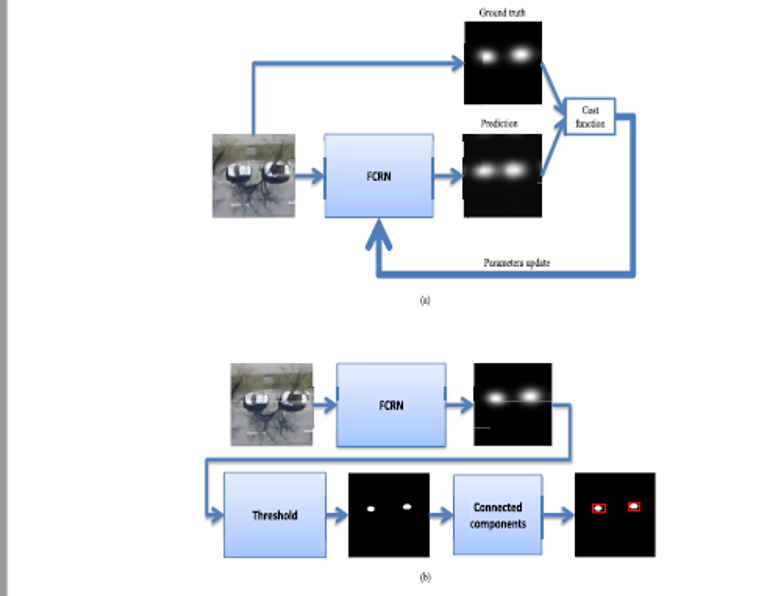}

\subsection{Feature-guided CNN for Denoising Images from Portable
Ultrasound Devices}
\label{sec:Feature}
\vspace{0.1cm}

Usually, the images captured by portable ultrasound equipment have considerable noise. For this reason, they proposed a novel denoising neural network model, called Feature guided Denoising Convolutional Neural Network (FDCNN), to remove noise while substantially retaining important feature information. In the training phase, they adopted a hierarchical noise addition strategy. The noise was added only in regions where no medical feature information was present. Finally, we combine this information with the denoising network. Experimental results show that the proposed medical image feature extraction method outperforms previous methods.Combined with the new denoising neural network architecture,portable ultrasound devices can now achieve better diagnostic
performance.

\vspace{0.3cm}
\includegraphics[scale = 1.8]{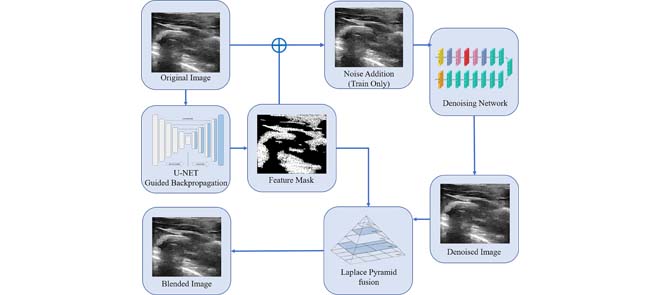}

\subsection{Assessing the Capability of Deep-Learning Models in
Parkinson’s Disease Diagnosis}
\label{sec:Disease}
\vspace{0.1cm}

This paper proposes and assess four deep-learning based models that classify patients based on biomarkers found in structural magnetic resonance images, and find that our 3DConvolution- Neural-Network model demonstrates high efficacy
in the task of diagnosing Parkinson’s disease, with an accuracy of 75 percent and 76 percentage sensitivity. Dataset used is obtained from the Parkinson’s Progression Markers Initiative (PPMI) public dataset. The best performing 3D
model was the 3D-CNN, with 75 percentage accuracy and balanced metrics in sensitivity, specificity and precision. It almost equally misdiagnoses with false-positives and falsenegatives. This paper uses an occlusion sensitivity analysis using best model, the 3DCNN. In the analysis, they allow the
model to fully train, and then obtain the test sample which maximally activates the PD-positive logit. We occlude a small 4 × 4 × 4 section of the sMRI sample by masking with zeros, and then obtain the new PD-positive logit value. This occlusion process is done for each block in the sample. We introduce a minimum threshold so that we can find the blocks which maximally activate the logit, which is necessary because all blocks activate it to some degree.

\vspace{0.3cm}
\includegraphics[scale = 0.5]{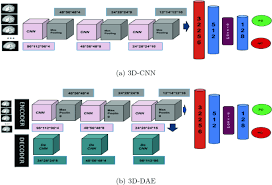}

\subsection{Edge-guided CNN for Denoising Images from Portable
Ultrasound Devices}
\label{sec:Edge}
\vspace{0.1cm}

To reduce long wait times and add convenience to patients,portable ultrasound scanning devices are becoming increasingly popular. These devices can be held in one hand, and are compatible with modern cell phones. However, the quality of
ultrasound images captured from the portable scanners is relatively poor compared to standard ultrasound scanning systems in hospitals.The proposed new neural network architecture called Edge-guided Denoising Convolutional Neural Network (EDCNN), which can preserve significant edge information in ultrasound images when removing noise. Involved edge guiding based on DnCNN and IRCNN and compared EDnCNN and E-IRCNN with the original DnCNN and IRCNN, showing significant improvement.

\vspace{0.3cm}
\includegraphics[scale = 0.5]{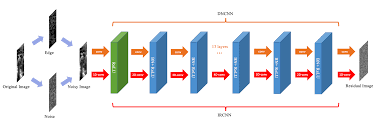}

\subsection{Robotic Catheter for Endovascular Surgery using 3D Magnetic
Guidance}
\label{sec:Endovasular}
\vspace{0.1cm}

Endovascular surgery is an alternative for invasive medical procedures that is becoming widely deployed for many procedures.One of the main challenges of this method is using X-ray before and during the surgery. X-ray side effects are
significant for surgeons who are regularly exposed while performing surgeries. A prototype system consisting of hardware and software parts has been designed. Two types of feedback were used for this system and one of them is servo sensor
and the other one is magnetic sensor feedback. Furthermore, the proposed system automatically guides the catheter through the simulated transparent vessels with different speeds and positions using the feedback from the Magnetic and servo sensors.

\section{Data}\label{sec:Roboflow}
The dataset used in this project was provided by the client consisting of cattle images captured by UAV's that is drones equipped with thermal cameras. This dataset was shared via Roboflow tool and consisted of 129 annotated thermal images of the cattle. These provided images are taken from different ariel angles so different cases such as half cut cattle images were included. We have split the given dataset as 103 Training set, 13 images for  Valid and Testing sets.

\vspace{0.3cm}
\includegraphics[scale = 0.45]{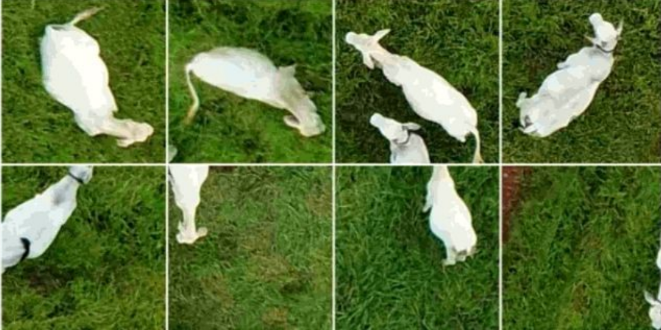}
\vspace{0.3cm}

Later we have also used the Roboflow tool to add a pre-processing element, Tiling and again converted the given dataset into 3x3 tiling which resulted in 1161 images which were divided as 927 images for training set, valid and test sets consisting 117 images each. 2x1 tiling resulted in a total of 258 images split as 206 images as train set and 26 each for valid and test sets. 

\vspace{0.3cm}
\includegraphics[scale = 0.45]{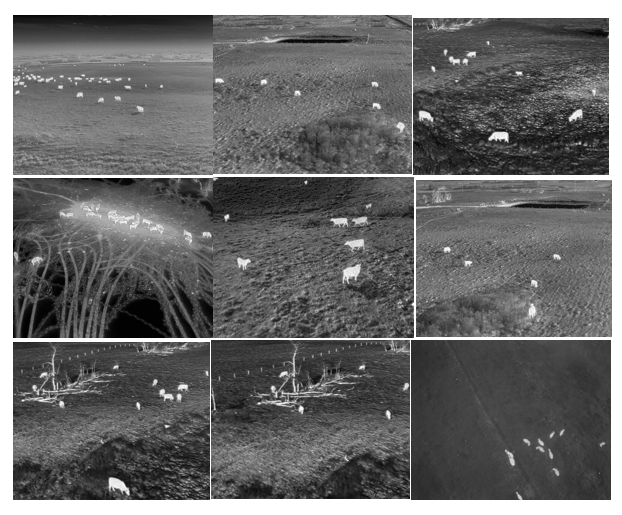}
\vspace{0.3cm}
\label{sec:dataset}
We have also used a RBG dataset from the internet provided by Harvard University for further testing consistsing of 249 images which were divided 179 images for training set, 49 images for validation set and 21 images for testing.

\vspace{0.3cm}
\includegraphics[scale = 0.3]{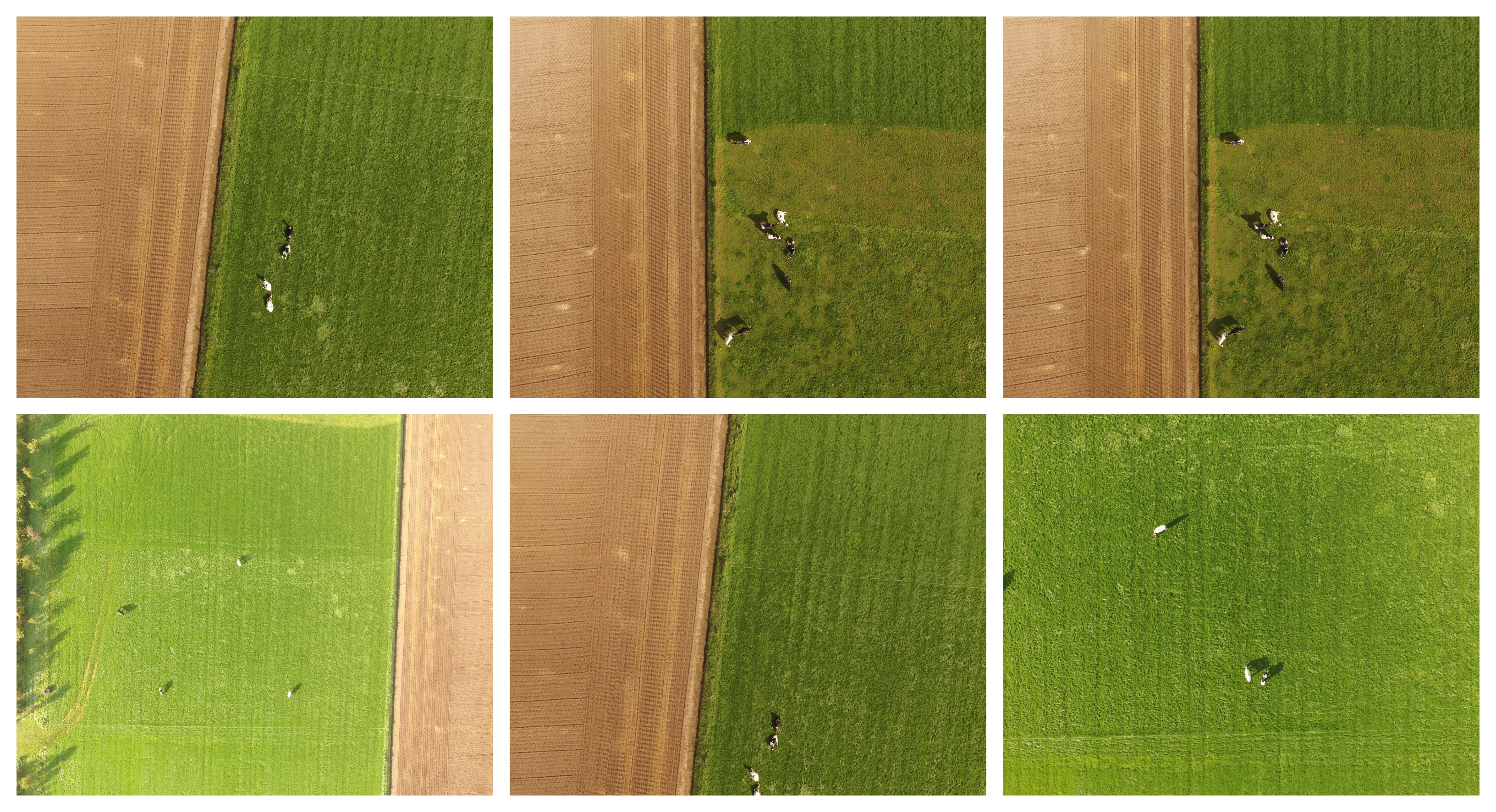}
\vspace{0.3cm}

\section{Methods}

Firstly we have used Yolov7 which uses Extended-ELAN (E-ELAN) to detect the cattle. We have trained the data as a batch of 16 for 55 epochs. We got a precision of 61.2 percentage. We could observe that the results were a bit more accurate when compared to other models.
\vspace{0.3cm}

\includegraphics[scale = 0.85]{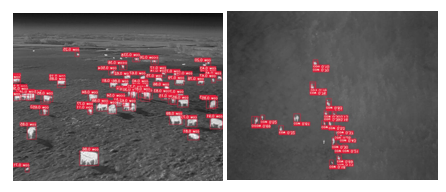}
Figure 3: showing detection results with Yolov7
\vspace{0.3cm}

We have used RetinaNet and made two models using EfficientNet and ResNet50 backbones. The results with ResNet50 backbone are more promising as compared to the EfficientNet backbone.In ResNet backbone, we have trained using 20 epochs and 100 steps. We got precision of about 58.3 percent.In EfficientNet backbone, we have used 100 epochs with batch size of 1 and precision is 50.4 percent.

\vspace{0.3cm}
\includegraphics[scale = 0.51]{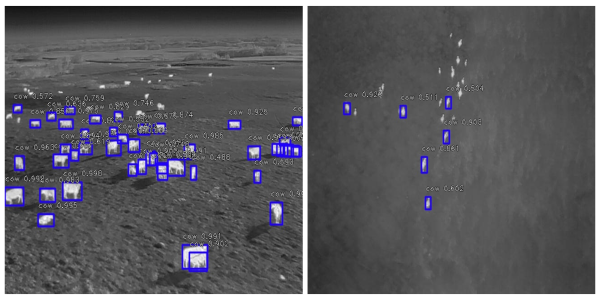}

Figure 3.1: showing detection results with RetinaNet-Resnet50
\vspace{0.3cm}

\includegraphics[scale = 0.45]{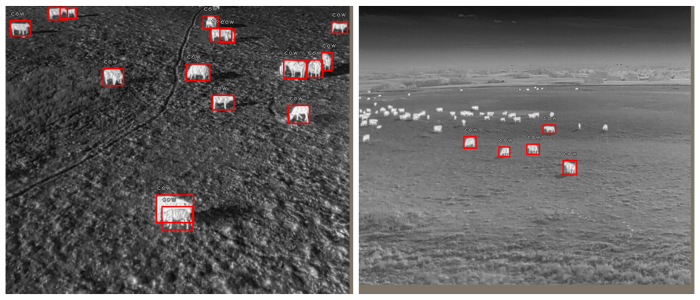}
Figure 3.2: showing detection results with RetinaNet-EfficientNet\\
\vspace{0.3cm}

Mask R-CNN was also used with backbone ResNet50 for detecting the cattle but the results was not that good when compared the YOLOv7 and RetinaNet.

\includegraphics[scale = 0.85]{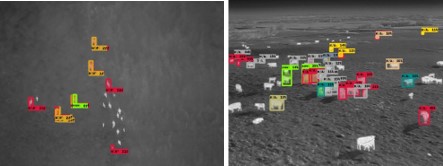}

\vspace{0.3cm}

We have used a pre-processing method Image Tiling as it is an important process for analysis of images with computer vision and allows for a more detailed look at specific sections of an image without sacrificing resolution. The technique is typically used for detecting small objects in high-resolution images.\\

We have used 2x1 Tiling and 3x3 Tiling on the dataset which divides each image into 2 rows and 1 column(for 2x1 Tiling) and for 3x3 tiling, each image is divided into 9 equal parts.

\section{Comparison Between Results in different case scenarios}

\subsection{Comparison for drone dataset}
 Below is the comparison between all four models(YOLOv7, RetinaNet with EfficientNet and ResNet50 as backbones and Mask RCNN) for the given dataset.
 
\vspace{0.3cm}
\includegraphics[scale = 0.45]{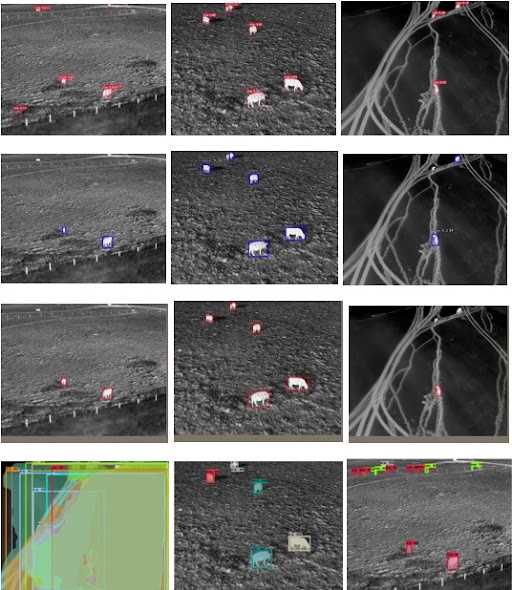}
Figure 4.0: showing results for all four models

\subsection{Comparison for 3x3 tiled dataset}
 Below is the comparison between all four models(YOLOv7, RetinaNet with EfficientNet and ResNet50 as backbones and Mask RCNN) for the 3x3 tiled dataset.
 
\vspace{0.3cm}
\includegraphics[scale = 0.45]{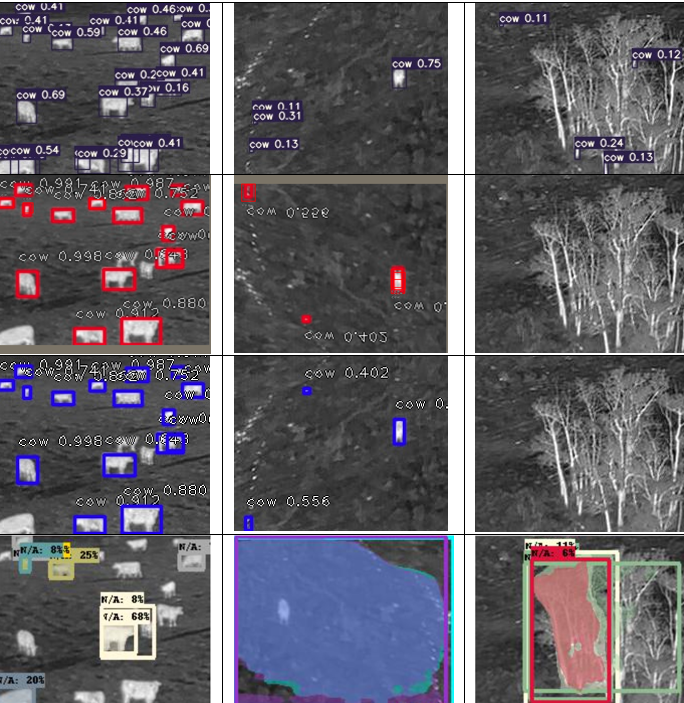}

Figure 4.1: showing 3X3 tiling results for all four models

\subsection{Comparison for 2x1 tiled dataset}
 Below is the comparison between all four models(Yolov7, RetinaNet with EfficientNet and ResNet50 as backbones and Mask RCNN) for the 2x1 tiled dataset.
 
\vspace{0.3cm}
\includegraphics[scale = 0.45]{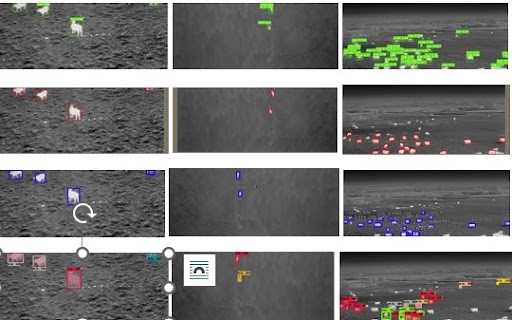}
Figure 4.2: showing 2x1 tiling results for all four models

\subsection{Comparison for RGB dataset}
 Below is the comparison between all four models(Yolov7, RetinaNet with EfficientNet and ResNet50 as backbones and Mask RCNN) for the RGB images.\\
 We can see in this case Mask RCNN performs better than other models and RetinaNet with EfficientNet performs worst.
 
\vspace{0.3cm}
\includegraphics[scale = 0.45]{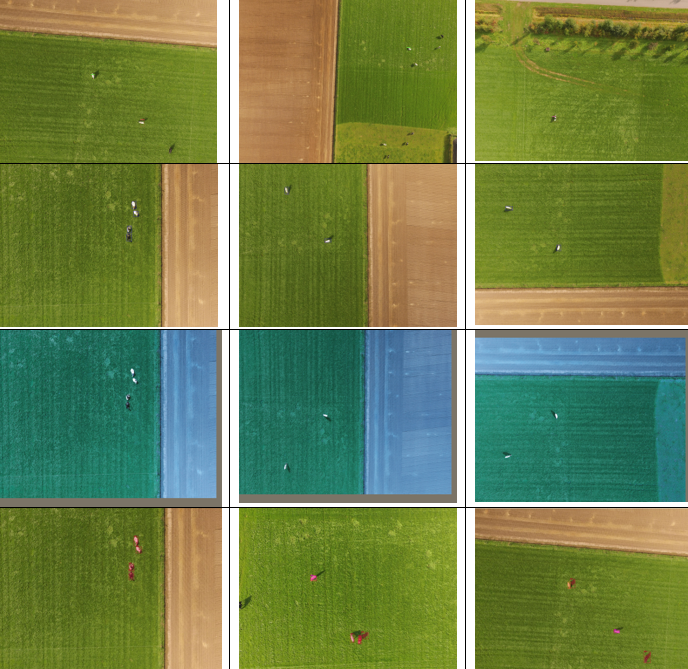}
Figure4.3 showing RGB images results for all four models

\section{Novelty in the project}
We have used two types of tiling to make detecting small images easy. We have used 3 by 3 tiling as well ans 2 by 1 tiling and tested these datasets with the four models.The 2 by 1 tiling gave better results than 3 by 3 tiling and out of all four models YOLOv7 gave more promising results in case of cattle detection and detection of hidden cattle i.e small objects.
Also the papers discussed in related works used ResNet,VGG and DenseNet as backbones with RetinaNet for detection of cattle. We have tried our RetinaNet model with EfficienNet backbone and results are quite good as it is able to detect small cattle easily.

\section{Conclusion}
Our main aim was to detect cattle through RGB and thermal images captured with help of drones. We tested four models in different scenarios, namely, YOLOv7,RetinaNet with ResNet50 backbone, RetinaNet with EfficientNet backbone and Mask RCNN.We also used 3 by 3 and 2 by 1 tiling to make hidden and small cattle detection possible.After Comparing results in all scenarios YoloV7 preformed better followed by Retinanet-ResNet50 and then RetinaNet-EffcientNet and Mask RCNN.
\section{Challenges and Future Works}
Our main challenge is to detect cattle which are not detected by any of the above four methods.To address the occlusion problem, we would like to focus on doing some research on advanced works done on small object detection and crowd counting that may give us some new ideas on cattle detection and stock monitoring in order to address the issue.  Then we will select some of the methods and test on our dataset to get started, and then improve the performance of the model by modifying the networks or adding new components to it.


\bibliographystyle{IEEEtran} 
\section{References}
\begin{enumerate}
   \item \hyperref[sec:vehicle]{Hilai et al., “Vehicle Detection and Counting in High-Resolution Aerial Images Using Convolutional Regression Neural Network”, IEEE Access Nov 16, 2017.}\\
   \item \hyperref[sec:astronomical] {Longmore et al., ”Adapting astronomical source detection software to help
detect animals in thermal images obtained by unmanned aerial systems”}\\
  \item \hyperref[sec:imbalance] {Benjamin et al., “Detecting Mammals in UAV Images: Best Practices to
address a substantially Imbalanced Dataset with Deep Learning”}\\
 \item \hyperref[sec:Retinex] {Changjiang et al., "Enhancement of low visibility aerial images using histogram truncation and an explicit Retinex representation for balancing contrast and color consistency"}\\
 \item \hyperref[sec:Edge] {Yingnan Ma et al., "Edge-guided CNN for Denoising Images from Portable Ultrasound Devices"}\\
 \item \hyperref[sec:Feature] {Guanfang Dong et al., "Feature-Guided CNN for Denoising Images From Portable Ultrasound Devices"}\\
  \item \hyperref[sec:oblique]{Jayme et al., “Cattle Detection Using Oblique UAV Images”},\\ 
  \item \hyperref[sec:LIDAR] {Jianfei et al., “Poisson Surface Reconstruction from LIDAR for Buttress Root Volume Estimation”}\\
   \item \hyperref[sec:Runway] {Harsh et al., “Synthetic Aerial Image Generation and Runway Segmentation”}\\
  \item \hyperref[sec:PCC] {Junyu et al., “PCC Net: Perspective Crowd Counting via Spatial Convolutional Network”}\\
  \item \hyperref[sec:zero] {Qiangqiang et al., “Dynamic Momentum Adaptation for Zero-Shot Cross-Domain Crowd Counting”}\\
   \item \hyperref[sec:freely]{Yingnan et al., “Background Subtraction based on Principal Motion for a Freely Moving Camera”.}\\
    \item \hyperref[sec:multiview]{Beibei Xu  et al., “Evaluation of Deep Learning for Automatic Multi-View Face Detection in Cattle}\\
   \item \hyperref[sec:choose] {Qingyu et al., "To Choose or to Fuse? Scale Selection for Crowd Counting”}\\
 \item \hyperref[sec:benchmark] {Longyin et al.“Detection, Tracking, and Counting Meets Drones in Crowds: A Benchmark”}\\
\item \hyperref[sec:zoomCount] {Usman et al.“ZoomCount: A Zooming Mechanism for Crowd Counting in Static Images”}\\
 \item \hyperref[sec:dehaze] {Yibin et al., “A multi-scale attentive recurrent network for image dehazing”, }\\
\item \hyperref[sec:fog] { Xinggang et al., “Fog Removal of Aerial Image Based on Gamma Correction and Guided Filtering”}\\
\item{ Wang et al., “YOLOv7: Trainable bag-of-freebies sets new state-of-the-art for real-time object detectors”}\\
\item{Tsung-Yi Lin et al., “Focal Loss for Dense Object Detection”}\\
 \item{https://developers.arcgis.com/python/guide/how-retinanet-works/}\\
 \item\hyperref[sec:Roboflow]{https://universe.roboflow.com/new-workspace-gr640/thermal\_cow.}\\
 \item\hyperref[sec:dataset]{https://dataverse.harvard.edu/file.xhtml?fileId=4771476
\&version=1.0}
\item{https://colab.research.google.com/drive/1ie
\_6DMnms6TzUjumIkVSL9J0fXiciGv3\#scrollTo=ycnYvdBXhedA}\\
 \item{ https://github.com/ViswanathaReddyGajjala/EfficientNet-RetinaNet}\\

\end{enumerate}

\end{document}